\documentclass[10pt,twocolumn,letterpaper]{article}

\usepackage[pagenumbers]{cvpr}

\usepackage{amsmath,amssymb,amsfonts}
\usepackage{graphicx}
\usepackage{booktabs}
\usepackage{multirow}
\usepackage{xcolor}
\usepackage{colortbl}
\usepackage{pifont}

\definecolor{officegrey}{HTML}{EDEDED}
\definecolor{officeorange}{HTML}{FBE5D6}

\definecolor{cvprblue}{rgb}{0.21,0.49,0.74}
\usepackage{hyperref}
\hypersetup{
    pagebackref,
    breaklinks,
    colorlinks,
    allcolors=cvprblue
}

\def\confName{CVPR}
\def\confYear{2027}

\title{Where to Search? Generative Search Center Prediction Enhancing \\Visual Object Tracking Under Fast Motion}

\author{\normalfont
Jiazheng Song$^{1}$ \quad
Rui Zhou$^{2}$ \quad
Ziqi Guo$^{1}$ \quad
Haining Zhang$^{1}$ \quad
Zhongqun Zhang$^{1}$ \\[4pt]
$^{1}$Nankai University \quad
$^{2}$University of Science and Technology of China
}

\begin{document}

\maketitle

\begin{abstract}
Existing single-object trackers commonly rely on a local motion assumption, restricting target search to a fixed-size region centered at the previously estimated target position. While effective under moderate motion, this strategy becomes unreliable under fast target motion, as large inter-frame displacements may cause the target to move beyond the predefined search region. To address this limitation, we propose WhereTrack, a plug-and-play framework that dynamically generates and selects search centers to improve tracker robustness under fast motion. Specifically, WhereTrack first employs a trajectory-conditioned flow matching model to generate multiple candidate search centers, capturing diverse hypotheses of future target locations. Furthermore, a reinforcement learning-based policy integrates visual and motion features to select the most promising search center from these candidates. To address the scarcity of fast-motion sequences in existing tracking benchmarks, we introduce the Fast Motion Dataset (FMD), comprising 178 video sequences across 48 object categories, of which 81.5\% exhibit fast target motion. Extensive experiments with 8 representative trackers across 6 benchmarks demonstrate that WhereTrack consistently improves tracking performance, achieving AUC improvements of up to 7.28\% without requiring modifications or fine-tuning of the original trackers. 

\end{abstract}

\section{Introduction}

\begin{figure}[t]
\centering
\includegraphics[width=\columnwidth]{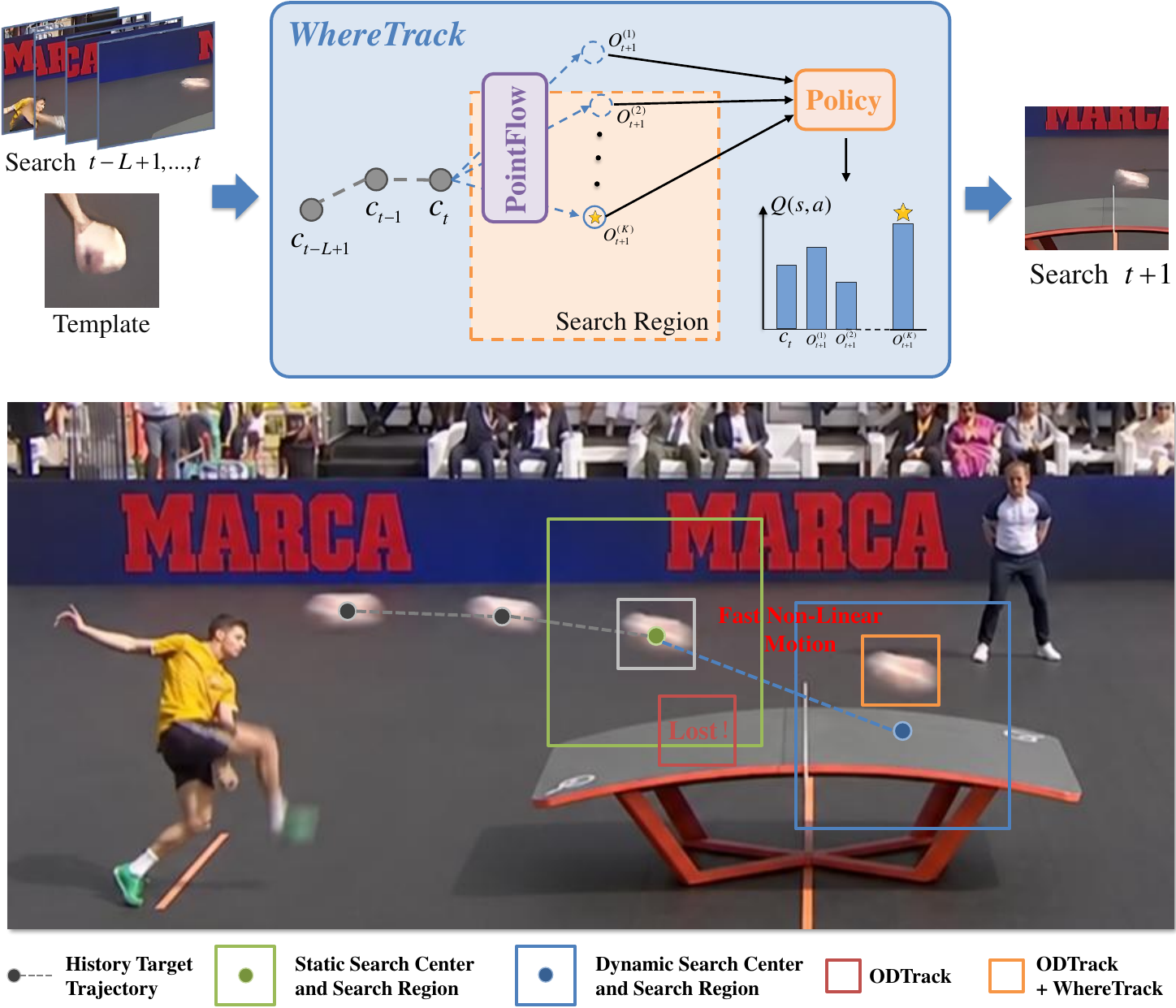}
\vspace{-6pt}
\caption{
\textbf{Motivation.} Under fast and non-linear motion, the ball moves outside the \textcolor[RGB]{151,184,82}{\textbf{static search region}}, causing \textcolor[RGB]{199,100,97}{\textbf{ODTrack}}~\cite{zheng2024odtrack} to lose the target. Our \textbf{WhereTrack} dynamically shifts the \textcolor[RGB]{73,124,187}{\textbf{dynamic search region}} to cover the target, enabling successful tracking with \textcolor[RGB]{247,148,67}{\textbf{ODTrack}}~\cite{zheng2024odtrack}. 
\textbf{Overview of our method.} Given historical target trajectories, PointFlow generates multiple candidate search centers, while Policy selects the most promising one using visual and motion cues to determine the next search region.
}
\label{fig:motivation}
\vspace{-8pt}
\end{figure}

Visual object tracking (VOT) is a fundamental computer vision task with widespread applications in autonomous driving~\cite{gao2019manifold}, aerial robotics~\cite{cao2021siamapn++}, and Embodied AI~\cite{liu2026trackvla++}. Given the target's initial bounding box, VOT aims to continuously estimate its location throughout a video sequence~\cite{wu2015object,fan2019lasot}. Most existing trackers follow a local matching paradigm, where a template is matched against a cropped search region to localize the target~\cite{bertinetto2016fully,li2019siamrpn++,ye2022joint}. During the past decade, VOT architectures have evolved from Siamese networks~\cite{bertinetto2016fully,chen2020siamban,bhat2019learning} to Transformer-based architectures~\cite{chen2021transformer,yan2021learning,cui2022mixformer,wu2023dropmae}, with recent advances incorporating cross-frame memory propagation~\cite{xie2023videotrack,cai2023hiptrack,zheng2024odtrack,spmtrack}, autoregressive modeling~\cite{chen2023seqtrack,wei2023autoregressive,xie2024autoregressive}, diffusion models~\cite{luo2024diffusiontrack}, and reinforcement learning-based bounding box refinement~\cite{relo}. Despite these advances, existing approaches focus primarily on improving feature representation, dynamic template~\cite{zheng2024odtrack}, and localization accuracy within predefined search regions.

However, these methods rely on a fundamental assumption that the target remains within the search window. As illustrated in Fig.~\ref{fig:motivation}, trackers such as ODTrack~\cite{zheng2024odtrack} typically center the next search window $o_{t+1}$ at the previously predicted target position ${c}_t$. Under fast or abrupt motion, large inter-frame displacements may move the target outside the search window, rendering subsequent feature matching ineffective.
Although motion models such as Kalman filters~\cite{kalman1960new} and LSTMs~\cite{shahbazi2018motion} can predict future target positions, their deterministic single-point predictions cannot represent multiple plausible future locations, making them less robust to sudden stops and sharp turns.

To address this limitation, we propose \textbf{WhereTrack}, a dynamic search center prediction framework that generates multiple candidate search centers and selects the most promising one for the next frame, as illustrated in Fig.~\ref{fig:motivation}. Specifically, we introduce PointFlow, a trajectory-conditioned flow matching model~\cite{lipman2023flow} that generates $K$ candidate search centers from historical target positions to capture diverse plausible future motions. Since motion history alone is insufficient to identify the most suitable search center, we further learn a Policy of Search Center Selection, an offline Q-learning-based network~\cite{watkins1992q,levine2020offline} that integrates visual and motion features to evaluate the generated candidates and a fallback option that retains the previous center. By adjusting only the search window position, WhereTrack can be seamlessly integrated into existing trackers without modifying their architectures or fine-tuning their parameters.

Furthermore, existing tracking benchmarks~\cite{wu2015object,fan2019lasot,huang2019got,muller2018trackingnet,fan2021lasot,wang2021towards} underrepresent fast-motion scenarios, limiting their ability to evaluate tracking failures caused by targets moving outside the search window. To address this gap, we introduce the Fast Motion Dataset (FMD), a benchmark specifically designed for extreme Fast Motion scenarios, comprising 178 video sequences and 49,395 frames across 48 object categories. FMD covers both sudden acceleration and sustained high-speed motion, with 81.5\% of its sequences exhibiting fast target motion, often accompanied by challenging attributes such as small targets (59.0\%), motion blur (56.7\%), and low resolution (52.8\%).

In this paper, we identify the limitations of static search windows under fast target motion and formulate search center positioning as a sequential decision problem that combines multi-hypothesis generation with visually guided selection. Our main contributions are summarized as follows:
\begin{itemize}
    \item We propose \textbf{PointFlow}, a trajectory-conditioned flow matching model for search center generation. By modeling the distribution of future target locations from historical trajectories, PointFlow generates diverse candidate search centers to account for motion uncertainty.
    \item We develop \textbf{Policy}, an offline reinforcement learning-based network for search center selection. By integrating visual and motion cues, Policy selects the most promising candidate search center. Together, PointFlow and Policy constitute WhereTrack, a plug-and-play framework that can be integrated into existing trackers without architectural modifications or fine-tuning.
    \item We introduce the \textbf{Fast Motion Dataset (FMD)}, comprising 178 video sequences and 49,395 frames across 48 object categories. With 81.5\% of its sequences exhibiting fast motion and annotations for eight challenging attributes, FMD enables a more comprehensive evaluation of tracking robustness under fast motion.

    \end{itemize}

\begin{figure*}[t]
\centering
\includegraphics[width=0.99\textwidth]{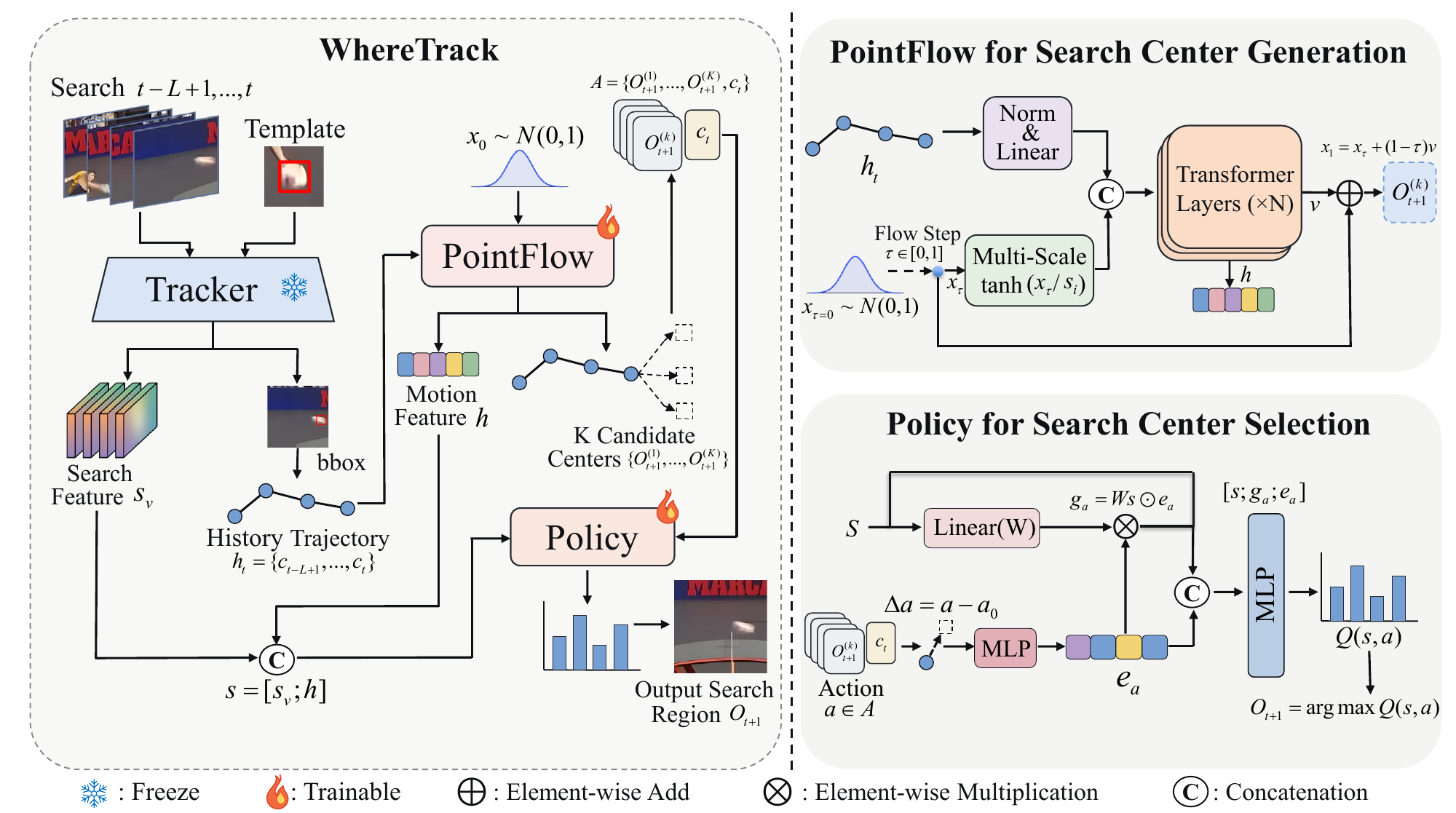}
\vspace{-8pt}
\caption{Overview of the WhereTrack framework. \textbf{Left (Main Pipeline):} Given the search sequence and template, the frozen tracker extracts the search feature $s_v$ and target center $c_t$. PointFlow generates $K$ candidate centers $\{O_{t+1}^{(1)}, \dots, O_{t+1}^{(K)}\}$ and motion feature $h$ from historical trajectory $h_t$, while Policy fuses state $s = [s_v; h]$ to evaluate the candidate set $A$ and select the next window center $O_{t+1}$. \textbf{Right:} Architectural details of the conditional flow matching proposer PointFlow and the offline Q-learning evaluator Policy.}
\label{fig:overview}
\vspace{-6pt}
\end{figure*}
\section{Related Work}

\subsection{Local Search Region-based Tracking}
Existing trackers commonly follow a local matching paradigm, extracting a search region centered at the previously estimated target position. Early Siamese-based trackers~\cite{bertinetto2016fully,li2019siamrpn++,chen2020siamban} and discriminative trackers~\cite{bhat2019learning} established effective local matching frameworks. Subsequent advances have primarily focused on improving target localization accuracy within the predefined search region. For instance, Transformer-based architectures~\cite{chen2021transformer,yan2021learning,ye2022joint,chen2022backbone,cui2022mixformer,wu2023dropmae,lin2024lorat} enhance feature interaction, while memory-based methods~\cite{xie2023videotrack,cai2023hiptrack,zheng2024odtrack,spmtrack} and motion-aware prompting~\cite{empl2025} improve temporal modeling. Other approaches explore autoregressive localization~\cite{chen2023seqtrack,wei2023autoregressive,xie2024autoregressive}, generative modeling~\cite{luo2024diffusiontrack,dreamtrack,tgtrack}, and reinforcement learning~\cite{relo} for target prediction and refinement. 

Despite these advances, fixed search regions may fail to cover the target under large inter-frame displacements. Unlike prior methods that improve localization within the search region, WhereTrack dynamically adjusts its center before image cropping.

\subsection{Search Region Adaptation for Tracking}
Existing search region adaptation methods primarily adjust either the search range or the search center. DaSiamRPN~\cite{zhu2018distractor} and SRRT~\cite{srrt} expand or resize the search region, while GlobalTrack~\cite{huang2020globaltrack} performs full-image search. However, enlarging the search region may introduce additional background distractors and computational overhead. Alternatively, motion-based approaches~\cite{shahbazi2018motion} employ recurrent neural networks (LSTMs) to predict the next search center from historical target trajectories. Nevertheless, such deterministic single-point predictions cannot represent multiple plausible future locations under abrupt motion changes. In contrast, WhereTrack combines generative multi-hypothesis prediction with visually guided policy-based selection to adapt the search center without enlarging the search region.

\subsection{Tracking Benchmarks for Fast Motion}
Evaluating tracking robustness under fast motion requires benchmarks with sufficient coverage of large inter-frame target displacements. Existing general-purpose benchmarks, including OTB100~\cite{wu2015object}, NUS\_PRO~\cite{li2016nus}, VOT~\cite{kristan2019visual}, TrackingNet~\cite{muller2018trackingnet}, GOT-10k~\cite{huang2019got}, LaSOT~\cite{fan2019lasot}, LaSOT-ext~\cite{fan2021lasot}, and TNL2K~\cite{wang2021towards}, cover diverse tracking scenarios but provide limited coverage of extreme fast motion. UAV benchmarks, such as UAV123~\cite{mueller2016benchmark}, UAV20L~\cite{mueller2016benchmark}, DTB70~\cite{li2017visual}, and UAVDT~\cite{du2018unmanned}, primarily focus on aerial-view tracking. Meanwhile, the high-frame-rate benchmark NfS~\cite{galoogahi2017need} provides videos captured at 240 fps, resulting in smaller inter-frame displacements when evaluated at the native frame rate. 

To complement these benchmarks, we construct the Fast Motion Dataset (FMD) by combining newly collected and annotated videos with curated sequences from public datasets. FMD comprises 178 sequences across 48 object categories, with 81.5\% exhibiting fast motion.


\section{Method}
As illustrated in Fig.~\ref{fig:overview} (left), WhereTrack consists of three components: a frozen base tracker, a trajectory-conditioned flow matching model (PointFlow, $G$), and a policy network (Policy, $Q$). At frame $t$, the base tracker predicts the target bounding box $b_t$ with center $c_t$ and extracts a target-aligned visual feature $s_v$. PointFlow generates multiple candidate search centers for the next frame from historical target positions, while Policy selects the most promising candidate based on visual and motion features. The selected center determines the next search window position without modifying the base tracker.


\subsection{Problem Formulation}
In conventional local-search tracking, the search window at frame $t+1$ is typically centered at the target position $c_t$ estimated in frame $t$:
\begin{equation}
O_{t+1} = c_t,
\label{eq:static_center}
\end{equation}
where $O$ means the center of search window.
Under fast or abrupt motion, however, large inter-frame displacements may place the target outside this search region.

WhereTrack instead formulates search center positioning as a two-stage process of candidate generation and policy-based selection. Given the historical target trajectory $h_t = {c_{t-L+1}, \ldots, c_t}$, PointFlow models a conditional distribution over future target locations and generates $K$ candidate search centers:
\begin{equation}
\{O_{t+1}^{(k)}\}_{k=1}^{K} \sim G(h_t),
\label{eq:proposer}
\end{equation}
where $G(x)$ means the PointFlow Network.
These candidates, together with the default center $a_0 = c_t$,
form an action set
$A_t = \{O_{t+1}^{(1)}, \ldots, O_{t+1}^{(K)}, c_t\}$. The Policy combines the target-aligned visual feature $s_v$ and historical motion feature $h$ into a state $s_t = [s_v; h]$, and selects the candidate with the highest estimated action value:
\begin{equation}
O_{t+1} = \underset{a \in A_t}{\arg\max} Q(s_t,a),
\label{eq:q_argmax}
\end{equation}
where $Q(x)$ means the Policy network.

To avoid unnecessary search window shifts, a conservative switching threshold $\delta$ is further introduced to regulate the selection process (Sec.~\ref{subsec:policy}). The selected $O_{t+1}$ determines only the cropping center for the next frame, leaving the base tracker's feature matching and bounding box regression unchanged.

\subsection{PointFlow for Search Center Generation}
\label{subsec:pointflow}
As illustrated in Fig.~\ref{fig:overview} (top right), PointFlow employs trajectory normalization and multi-scale displacement encoding to accommodate varying target scales and motion magnitudes. Given the historical trajectory $h_t = \{c_{t-L+1}, \ldots, c_t\}$, we normalize the coordinates with respect to the current target center $c_t$ and the search window size, followed by a linear projection to obtain trajectory features. During training, we sample Gaussian noise $x_0 \sim \mathcal{N}(\mathbf{0}, I)$ and interpolate it with the normalized ground-truth displacement $x_{\mathrm{gt}}$ at flow time $\tau \in [0,1]$:
\[
x_\tau = (1-\tau)x_0 + \tau x_{\mathrm{gt}}.
\]
To capture motion across different scales, we encode $x_\tau$ using multiple scale factors $\{s_i\}$, yielding multi-scale displacement features $\{\tanh(x_\tau/s_i)\}$. These features are concatenated with the trajectory features and processed by $N$ Transformer layers to predict the velocity field $v$, while the penultimate layer provides a historical motion representation $h$.

We train PointFlow to regress the ground-truth velocity $v_{\mathrm{gt}} = x_{\mathrm{gt}} - x_0$. To improve endpoint accuracy with fewer integration steps, we further estimate the target displacement as
\begin{equation}
    \hat{x}_1 = x_\tau + (1-\tau)v,
    \label{eq:x1_hat}
\end{equation}
and optimize the following objective:
\begin{equation}
    \mathcal{L}_{\mathrm{flow}} =
    \|v-v_{\mathrm{gt}}\|^2
    + \lambda\|\hat{x}_1-x_{\mathrm{gt}}\|^2,
    \label{eq:loss_cfm}
\end{equation}
where $\lambda$ controls the contribution of endpoint reconstruction. During inference, the predicted displacements are transformed into image coordinates to obtain $K$ candidate search centers $\{O_{t+1}^{(k)}\}_{k=1}^{K}$. To improve robustness to noisy tracker predictions, we augment historical trajectories with random positional offsets and search window scale perturbations during training.

\subsection{Policy for Search Center Selection}
\label{subsec:policy}

We introduce the Policy module, an offline Q-learning-based network~\cite{watkins1992q,levine2020offline} that evaluates candidate search centers by integrating visual and motion features. The Policy module consists of three key components: state and action representation for candidate evaluation, offline policy learning with two-step returns, and conservative search center selection to avoid unnecessary window shifts. The overall architecture is illustrated in Fig.~\ref{fig:overview} (bottom right).

\textbf{State and Action Representation.}
To capture target-specific visual cues while suppressing background interference, we apply ROI-Align~\cite{he2017mask} to extract a $7 \times 7$ feature map from the predicted bounding box $b_t$, which is projected into a visual feature $s_v$. This feature is concatenated with the historical motion representation $h$ to form the state $s_t = [s_v; h]$. For each candidate action $a \in A_t$, we encode its normalized displacement relative to the default center $a_0 = c_t$ using an MLP, yielding an action embedding $e_a$. To model state–action compatibility, we compute an interaction feature $g_a = W s_t \odot e_a$, where $W$ denotes a learnable linear projection and $\odot$ is element-wise multiplication. The concatenated representation $[s_t; g_a; e_a]$ is then fed into a shared three-layer MLP to predict the action value $Q(s_t,a)$ for each candidate.

\textbf{Offline Policy Learning.}
We collect training samples through two-step offline rollouts, using the IoU gain over the default action as the reward, $r_t(a)=\mathrm{IoU}_t(a)-\mathrm{IoU}_t(a_0)$. The two-step return target is defined as
\begin{equation}
    y_t(a) = r_t(a) + \gamma \max_{a' \in A_{t+1}} r_{t+1}(a'),
    \label{eq:bellman_target}
\end{equation}
where $\gamma$ is the discount factor and $A_{t+1}$ is obtained by rolling out the tracker after action $a$. We optimize Policy with
\begin{equation}
    \mathcal{L}_{\mathrm{policy}} =
    \big(Q(s_t,a)-y_t(a)\big)^2.
    \label{eq:loss_policy}
\end{equation}

\textbf{Conservative Search Center Selection.}
During inference, Policy selects the candidate with the highest Q-value among the $K$ generated search centers, denoted as $O_{t+1}^{(m)}$. To avoid unnecessary window shifts under smooth motion, we introduce a conservative threshold $\delta > 0$:
\begin{equation}
    O_{t+1} = 
    \begin{cases} 
        O_{t+1}^{(m)}, & \text{if } Q(s, O_{t+1}^{(m)}) - Q(s, a_0) > \delta, \\
        c_t, & \text{otherwise.}
    \end{cases}
    \label{eq:gate_rule}
\end{equation}

\section{The FMD Dataset}

\noindent\textbf{Tracking sequences.} 
To evaluate tracking robustness under fast motion, particularly when targets move beyond predefined search regions, we construct the Fast Motion Dataset (FMD), comprising 178 video sequences across 48 object categories. The videos cover resolutions from $640 \times 360$ to $3840 \times 2160$, with frame rates ranging from 23.98 to 50 fps.

\begin{table}[t]
\centering
\footnotesize
\caption{Comparison of FMD with mainstream tracking benchmarks in sequence count, target classes, and frame length statistics. Classes for OTB100, UAV123, UAV20L,
and DTB70 are counts of distinct target types obtained by merging sequence-name variants. A dash (--) indicates that the benchmark does not provide a class
taxonomy (TNL2K).}
\label{tab:dataset_comparison}
\setlength{\tabcolsep}{5pt}
\begin{tabular}{lcccccc}
\toprule
Dataset & Year & Seqs & Classes & Min & Max & Total \\
\midrule
OTB100~\cite{wu2015object} & 2015 & 100 & 69 & 71 & 3,872 & 59K \\
NUS\_PRO~\cite{li2016nus} & 2016 & 365 & 17 & 146 & 5,040 & 135K \\
UAV123~\cite{mueller2016benchmark} & 2016 & 123 & 10 & 109 & 3,085 & 113K \\
UAV20L~\cite{mueller2016benchmark} & 2016 & 20 & 6 & 1,717 & 5,527 & 59K \\
DTB70~\cite{li2017visual} & 2017 & 70 & 29 & 68 & 699 & 16K \\
NfS~\cite{galoogahi2017need} & 2017 & 100 & 17 & 169 & 20,665 & 383K \\
TrackingNet~\cite{muller2018trackingnet} & 2018 & 511 & 27 & 96 & 2,368 & 226K \\
GOT-10k~\cite{huang2019got} & 2019 & 180 & 84 & 51 & 920 & 23K \\
LaSOT~\cite{fan2019lasot} & 2019 & 280 & 70 & 1,000 & 9,999 & 685K \\
LaSOT-ext~\cite{fan2021lasot} & 2021 & 150 & 15 & 1,000 & 5,999 & 375K \\
DepthTrack~\cite{yan2021depthtrack} & 2021 & 50 & 40 & 146 & 4,095 & 76K \\
TNL2K~\cite{wang2021towards} & 2021 & 700 & -- & 21 & 18,488 & 435K \\
\midrule
\rowcolor{officeorange!55}
\textbf{FMD (Ours)} & \textbf{2026} & \textbf{178} & \textbf{48} & \textbf{45} & \textbf{2,500} & \textbf{49K} \\
\bottomrule
\end{tabular}
\vspace{-6pt}
\end{table}

As summarized in Table~\ref{tab:dataset_comparison}, FMD comprises 178 sequences and 49,395 frames across 48 object categories, with sequence lengths ranging from 45 to 2,500 frames. To ensure motion and category diversity, we construct FMD from two sources: 49 sequences (16,065 frames) collected from 21 YouTube videos and 129 non-overlapping fast-motion segments (33,330 frames) curated from 90 public videos, including GOT-10k~\cite{huang2019got}, LaSOT~\cite{fan2019lasot}, LaSOT-ext~\cite{fan2021lasot}, and NUS\_PRO~\cite{li2016nus}. The two subsets cover 21 and 28 object categories, respectively, sharing only the volleyball category.


\noindent\textbf{Data annotation.} 
FMD provides frame-level bounding box annotations and sequence-level labels for eight challenging attributes. The 49 newly collected sequences were manually annotated using CVAT~\cite{cvat}, while the 129 sequences curated from public datasets retain their original annotations~\cite{huang2019got,fan2019lasot,fan2021lasot,li2016nus}. Fully occluded or out-of-view targets are uniformly labeled as NaN.

The eight attributes include Fast Motion (FM), Scale Variation (SV), Aspect Ratio Change (ARC), Small Target (ST), Low Resolution (LR), Occlusion (OCC), Motion Blur (MB), and Camera Motion (CM). FM is defined as a target displacement exceeding one target length in at least 10\% of frames, while SV, ARC, ST, LR, and OCC follow established definitions~\cite{huang2019got,mueller2016benchmark}. MB and CM are automatically annotated using target-region sharpness and inter-frame background changes, respectively. We provide more details about dataset in the supplementary material.
\begin{figure}[!t]
\centering
\includegraphics[width=0.98\columnwidth]{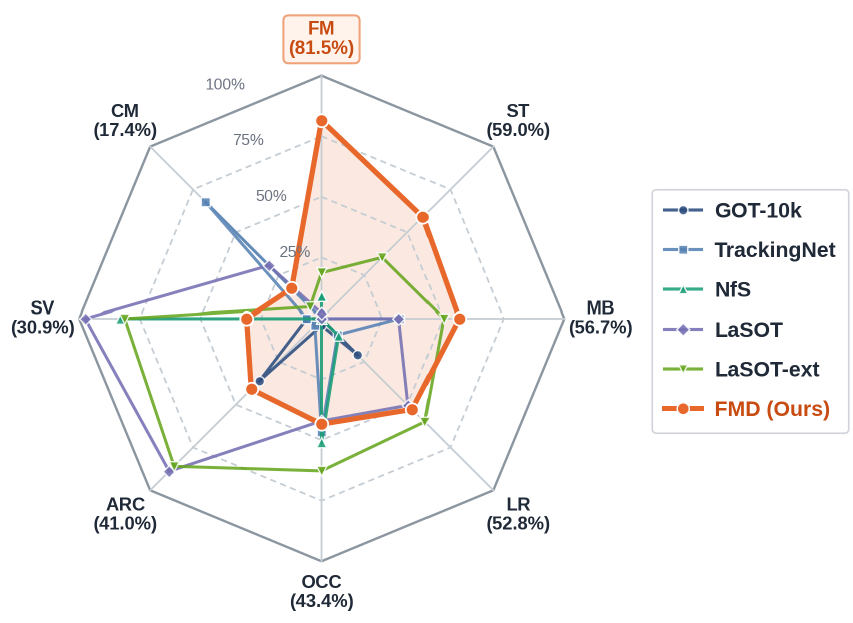}
\vspace{-8pt}
\caption{Radar chart comparing the percentages of sequences exhibiting 8 challenging attributes across FMD and other 5 widely used benchmarks.
For consistent comparison, FM is defined as a frame-to-frame target displacement exceeding one target body length in at least 10\% of frames within a sequence.
Notably, FMD exhibits a higher proportion of sequences with Fast Motion (FM), highlighting its stronger coverage of challenging rapid-motion scenarios.}
\label{fig:radar}
\vspace{-8pt}
\end{figure}
\section{Experiments}

We evaluate WhereTrack on the proposed FMD dataset and five mainstream single object tracking benchmarks: NfS~\cite{galoogahi2017need}, LaSOT~\cite{fan2019lasot}, LaSOT-ext~\cite{fan2021lasot}, TrackingNet~\cite{muller2018trackingnet}, and GOT-10k~\cite{huang2019got}. This section first describes the experimental setup, then reports quantitative comparisons across eight representative base trackers~\cite{zheng2024odtrack,ye2022joint,spmtrack,lin2024lorat,relo,chen2023seqtrack,mcitrack,sutrack}, attribute-specific analyses, and qualitative comparisons, and finally presents ablation studies and efficiency analyses based on ODTrack~\cite{zheng2024odtrack}.

\subsection{Experimental Setup}

\noindent\textbf{Training.}
We train PointFlow and Policy in two stages on four NVIDIA RTX 4090 GPUs while keeping all parameters of the base tracker frozen. The training set comprises the training splits of LaSOT~\cite{fan2019lasot}, GOT-10k~\cite{huang2019got}, and TrackingNet~\cite{muller2018trackingnet}. In the first stage, PointFlow is trained on ground-truth target center trajectories using Eq.~(\ref{eq:loss_cfm}), with the historical trajectory length set to $L=4$ and the endpoint reconstruction weight set to $\lambda=1.0$. To simulate cumulative localization drift during inference, random positional offsets and window scale perturbations are injected into the historical trajectories during training. In the second stage, tracking rollouts under different candidate windows are collected on training sequences via two-step offline expansion, and Policy is optimized using the mean squared error loss in Eq.~(\ref{eq:loss_policy}) with a discount factor of $\gamma=0.5$. Both modules are trained using AdamW~\cite{loshchilov2019decoupled}.

\begin{table*}[t]
\centering
\scriptsize
\caption{Quantitative comparison of eight representative base trackers equipped with different search region adaptation strategies (SRRT~\cite{srrt}, LSTM~\cite{shahbazi2018motion}, and our WhereTrack) across six tracking benchmarks. The best and second-best results within each tracker group are highlighted in \textbf{bold} and \underline{underlined}, respectively.}
\label{tab:main_results}
\setlength{\tabcolsep}{2.4pt}
\renewcommand{\arraystretch}{0.96}
\begin{tabular*}{\textwidth}{@{\extracolsep{\fill}}l ccc ccc ccc ccc ccc ccc}
\toprule
\multirow{2}{*}{Method}
& \multicolumn{3}{c}{FMD}
& \multicolumn{3}{c}{LaSOT}
& \multicolumn{3}{c}{LaSOT-ext}
& \multicolumn{3}{c}{TrackingNet}
& \multicolumn{3}{c}{GOT-10k}
& \multicolumn{3}{c}{NfS} \\
\cmidrule(lr){2-4}
\cmidrule(lr){5-7}
\cmidrule(lr){8-10}
\cmidrule(lr){11-13}
\cmidrule(lr){14-16}
\cmidrule(lr){17-19}
& AUC & $\mathrm{P}_{\mathrm{norm}}$ & P
& AUC & $\mathrm{P}_{\mathrm{norm}}$ & P
& AUC & $\mathrm{P}_{\mathrm{norm}}$ & P
& AUC & $\mathrm{P}_{\mathrm{norm}}$ & P
& AO & $\mathrm{SR}_{0.5}$ & $\mathrm{SR}_{0.75}$
& AUC & $\mathrm{P}_{\mathrm{norm}}$ & P \\
\midrule

ODTrack~\cite{zheng2024odtrack}
& 54.93 & 66.70 & 72.44
& 72.74 & 82.63 & 80.05
& \underline{50.73} & \underline{61.85} & \underline{58.19}
& 85.10 & 90.14 & 84.88
& \underline{0.7752} & \underline{0.8680} & \underline{0.7612}
& 67.38 & 85.00 & 84.46 \\

+SRRT~\cite{srrt}
& 52.72 & 64.67 & 70.88
& 70.25 & 81.40 & 77.27
& 46.89 & 57.84 & 53.89
& 80.07 & 87.81 & 80.28
& 0.7404 & 0.8455 & 0.7198
& 64.19 & 84.09 & 82.17 \\

+LSTM~\cite{shahbazi2018motion}
& \underline{57.51} & \underline{69.86} & \underline{75.56}
& \underline{72.84} & \underline{82.76} & \underline{80.07}
& 50.12 & 61.00 & 57.31
& \textbf{85.16} & \textbf{90.22} & \textbf{84.95}
& 0.7687 & 0.8588 & 0.7558
& \underline{67.68} & \underline{85.43} & \underline{84.72} \\

\rowcolor{officegrey}
+WhereTrack
& \textbf{61.22} & \textbf{74.40} & \textbf{80.94}
& \textbf{73.38} & \textbf{83.40} & \textbf{80.81}
& \textbf{52.50} & \textbf{63.95} & \textbf{60.45}
& \underline{85.14} & \underline{90.21} & \underline{84.93}
& \textbf{0.7766} & \textbf{0.8732} & \textbf{0.7690}
& \textbf{68.70} & \textbf{86.89} & \textbf{86.15} \\

\midrule

OSTrack~\cite{ye2022joint}
& 52.96 & 64.15 & 69.91
& 70.79 & 80.73 & 77.19
& 50.54 & 61.25 & 57.64
& \underline{83.74} & \underline{88.32} & \underline{83.01}
& 0.7440 & 0.8393 & 0.7244
& 67.46 & 83.93 & 82.82 \\

+SRRT~\cite{srrt}
& 50.61 & 61.82 & 68.41
& 66.25 & 77.01 & 71.99
& 47.98 & 59.18 & 55.51
& 79.17 & 86.05 & 78.20
& 0.6926 & 0.7917 & 0.6413
& 64.63 & 82.54 & 80.13 \\

+LSTM~\cite{shahbazi2018motion}
& \underline{56.42} & \underline{68.51} & \underline{74.76}
& \underline{71.12} & \underline{81.09} & \underline{77.52}
& \underline{50.80} & \underline{61.54} & \underline{57.96}
& 83.73 & 88.31 & 83.00
& \textbf{0.7532} & \textbf{0.8517} & \textbf{0.7370}
& \textbf{68.30} & \underline{85.02} & \underline{83.89} \\

\rowcolor{officegrey}
+WhereTrack
& \textbf{56.85} & \textbf{69.16} & \textbf{75.41}
& \textbf{71.50} & \textbf{81.60} & \textbf{78.12}
& \textbf{51.40} & \textbf{62.28} & \textbf{58.74}
& \textbf{83.88} & \textbf{88.51} & \textbf{83.21}
& \underline{0.7484} & \underline{0.8448} & \underline{0.7300}
& \underline{68.26} & \textbf{85.15} & \textbf{83.93} \\

\midrule

SPMTrack~\cite{spmtrack}
& 50.96 & 62.29 & 67.30
& \underline{76.81} & \underline{85.90} & \underline{83.97}
& 52.91 & 64.22 & 60.20
& \underline{86.67} & \underline{91.41} & \underline{87.58}
& \underline{0.8191} & \underline{0.9116} & 0.8265
& \underline{70.50} & 87.49 & \underline{86.61} \\

+SRRT~\cite{srrt}
& 36.69 & 46.71 & 52.94
& 37.97 & 45.16 & 43.33
& 22.81 & 29.91 & 27.74
& 33.18 & 55.40 & 39.20
& 0.4393 & 0.4346 & 0.3104
& 38.64 & 53.92 & 55.12 \\

+LSTM~\cite{shahbazi2018motion}
& \underline{52.65} & \underline{64.30} & \underline{69.57}
& 74.48 & 84.56 & 83.46
& \underline{53.86} & \underline{65.48} & \underline{61.60}
& \underline{86.67} & \textbf{91.42} & \textbf{87.61}
& \textbf{0.8200} & \textbf{0.9134} & \textbf{0.8296}
& 69.96 & \underline{87.59} & 86.54 \\

\rowcolor{officegrey}
+WhereTrack
& \textbf{56.97} & \textbf{69.76} & \textbf{75.64}
& \textbf{77.23} & \textbf{86.05} & \textbf{83.98}
& \textbf{55.88} & \textbf{67.89} & \textbf{64.19}
& \textbf{86.68} & 91.40 & \underline{87.58}
& 0.8138 & 0.9032 & \underline{0.8286}
& \textbf{70.73} & \textbf{87.78} & \textbf{86.71} \\

\midrule

LoRAT~\cite{lin2024lorat}
& 54.33 & 66.44 & 72.08
& \underline{73.50} & \underline{83.53} & \underline{82.36}
& 55.77 & 68.08 & 64.35
& \underline{85.27} & \underline{90.11} & \underline{85.69}
& 0.7932 & 0.8841 & 0.8054
& \underline{69.31} & \underline{86.19} & \underline{84.60} \\

+SRRT~\cite{srrt}
& 34.41 & 42.21 & 48.39
& 39.63 & 46.76 & 44.01
& 25.68 & 34.03 & 30.88
& 41.78 & 60.46 & 45.70
& 0.4795 & 0.4997 & 0.3504
& 41.68 & 56.73 & 55.32 \\

+LSTM~\cite{shahbazi2018motion}
& \underline{55.53} & \underline{68.10} & \underline{73.64}
& 73.45 & 83.49 & 82.29
& \underline{55.90} & \underline{68.19} & \underline{64.52}
& \underline{85.27} & 90.10 & \underline{85.69}
& \underline{0.7983} & \underline{0.8905} & \textbf{0.8097}
& 69.20 & 86.17 & 84.52 \\

\rowcolor{officegrey}
+WhereTrack
& \textbf{58.00} & \textbf{71.14} & \textbf{77.17}
& \textbf{73.98} & \textbf{84.13} & \textbf{82.94}
& \textbf{56.64} & \textbf{69.12} & \textbf{65.54}
& \textbf{85.29} & \textbf{90.14} & \textbf{85.71}
& \textbf{0.7996} & \textbf{0.8931} & \underline{0.8060}
& \textbf{69.49} & \textbf{86.52} & \textbf{85.07} \\

\midrule

RELO~\cite{relo}
& 53.46 & 63.65 & 72.15
& \underline{74.99} & \underline{84.88} & \underline{83.19}
& \underline{57.86} & \underline{69.65} & \underline{66.80}
& \underline{87.31} & \underline{91.68} & \underline{88.01}
& \textbf{0.8071} & 0.8933 & 0.8248
& 71.00 & \underline{87.44} & 86.23 \\

+SRRT~\cite{srrt}
& 48.36 & 57.85 & 64.72
& 72.32 & 81.93 & 79.84
& 51.34 & 61.65 & 58.12
& 85.14 & 90.34 & 86.24
& 0.7713 & 0.8582 & 0.7690
& 67.73 & 83.48 & 82.46 \\

+LSTM~\cite{shahbazi2018motion}
& \underline{54.75} & \underline{65.76} & \underline{73.69}
& 74.88 & 84.78 & 83.11
& 56.78 & 68.39 & 65.58
& 87.29 & 91.66 & \textbf{88.04}
& 0.8057 & \underline{0.8939} & \underline{0.8267}
& \underline{71.10} & 87.35 & \underline{86.58} \\

\rowcolor{officegrey}
+WhereTrack
& \textbf{57.36} & \textbf{68.86} & \textbf{77.19}
& \textbf{75.32} & \textbf{85.31} & \textbf{83.66}
& \textbf{58.11} & \textbf{69.93} & \textbf{67.34}
& \textbf{87.33} & \textbf{91.69} & \underline{88.01}
& \underline{0.8070} & \textbf{0.8967} & \textbf{0.8302}
& \textbf{71.66} & \textbf{88.21} & \textbf{87.18} \\

\midrule

SeqTrack~\cite{chen2023seqtrack}
& 40.87 & 49.10 & 54.13
& \underline{72.51} & \underline{81.53} & \underline{79.25}
& 50.66 & 61.55 & 57.48
& 85.48 & 89.85 & 85.81
& 0.7663 & 0.8383 & 0.7526
& 66.21 & 83.55 & 81.89 \\

+SRRT~\cite{srrt}
& \underline{44.81} & \underline{54.42} & \underline{61.06}
& 68.22 & 78.15 & 73.75
& 48.14 & 59.17 & 55.18
& 80.51 & 86.39 & 80.06
& 0.7261 & 0.8231 & 0.6811
& 63.35 & 79.20 & 77.74 \\

+LSTM~\cite{shahbazi2018motion}
& 43.90 & 52.95 & 58.11
& 71.89 & 80.77 & 78.56
& \underline{50.70} & \underline{61.57} & \underline{57.52}
& \underline{85.55} & \underline{89.90} & \underline{85.87}
& \textbf{0.7743} & \textbf{0.8476} & \textbf{0.7633}
& \underline{66.63} & \underline{84.13} & \underline{82.53} \\

\rowcolor{officegrey}
+WhereTrack
& \textbf{46.32} & \textbf{55.94} & \textbf{61.66}
& \textbf{73.14} & \textbf{82.28} & \textbf{79.99}
& \textbf{52.29} & \textbf{63.48} & \textbf{59.82}
& \textbf{85.57} & \textbf{89.92} & \textbf{85.88}
& \underline{0.7687} & \underline{0.8424} & \underline{0.7558}
& \textbf{67.22} & \textbf{84.90} & \textbf{83.23} \\

\midrule

MCITrack~\cite{mcitrack}
& 42.81 & 51.30 & 56.01
& \underline{76.62} & \underline{86.19} & \underline{85.30}
& \underline{54.54} & 64.99 & \underline{61.58}
& \underline{87.87} & \underline{92.07} & 89.19
& 0.8080 & 0.8878 & 0.8233
& 69.40 & 85.13 & 84.24 \\

+SRRT~\cite{srrt}
& \underline{49.57} & \underline{59.50} & \underline{64.10}
& 75.94 & 85.74 & 84.24
& 50.89 & 60.19 & 56.60
& 86.37 & 91.16 & 87.80
& 0.7910 & 0.8712 & 0.7944
& 67.74 & 82.94 & 82.04 \\

+LSTM~\cite{shahbazi2018motion}
& 47.88 & 57.98 & 63.01
& 76.49 & 85.93 & 84.90
& 54.52 & \underline{65.00} & 61.54
& 87.86 & 92.04 & \textbf{89.21}
& \textbf{0.8126} & \textbf{0.8922} & \textbf{0.8284}
& \underline{70.05} & \underline{85.83} & \underline{85.12} \\

\rowcolor{officegrey}
+WhereTrack
& \textbf{50.09} & \textbf{60.45} & \textbf{64.77}
& \textbf{77.33} & \textbf{86.93} & \textbf{86.05}
& \textbf{55.72} & \textbf{66.32} & \textbf{63.06}
& \textbf{87.88} & \textbf{92.09} & \textbf{89.21}
& \underline{0.8093} & \underline{0.8881} & \underline{0.8235}
& \textbf{70.95} & \textbf{87.05} & \textbf{86.21} \\

\midrule

SUTrack~\cite{sutrack}
& 46.14 & 55.45 & 60.66
& \underline{74.77} & \underline{84.31} & \underline{82.58}
& 53.49 & 63.88 & 60.65
& \underline{87.62} & \underline{91.66} & \underline{88.63}
& 0.8190 & 0.8981 & \underline{0.8399}
& 69.65 & 86.06 & 84.87 \\

+SRRT~\cite{srrt}
& \underline{48.76} & \underline{59.78} & \underline{65.38}
& 72.72 & 82.13 & 80.15
& 52.97 & 63.18 & 59.41
& 85.70 & 90.52 & 86.90
& 0.7889 & 0.8700 & 0.7949
& 69.09 & 85.98 & 84.93 \\

+LSTM~\cite{shahbazi2018motion}
& 47.93 & 57.82 & 62.78
& 74.19 & 83.80 & 82.20
& \underline{53.78} & \underline{64.10} & \underline{60.82}
& 87.56 & 91.59 & 88.57
& \underline{0.8193} & \underline{0.8990} & \textbf{0.8413}
& \underline{70.74} & \underline{87.46} & \underline{86.19} \\

\rowcolor{officegrey}
+WhereTrack
& \textbf{50.03} & \textbf{60.74} & \textbf{66.29}
& \textbf{75.63} & \textbf{85.52} & \textbf{84.06}
& \textbf{55.77} & \textbf{66.74} & \textbf{63.55}
& \textbf{87.63} & \textbf{91.68} & \textbf{88.65}
& \textbf{0.8231} & \textbf{0.9042} & \underline{0.8399}
& \textbf{71.08} & \textbf{88.10} & \textbf{86.71} \\

\bottomrule
\end{tabular*}
\vspace{-4pt}
\end{table*}

\noindent\textbf{Inference and baselines.}
To evaluate the generalization ability of WhereTrack across diverse tracking paradigms, we directly integrate the trained PointFlow and Policy into eight open-source base trackers without re-training, covering one-stream and unified Transformer trackers (OSTrack~\cite{ye2022joint}, LoRAT~\cite{lin2024lorat}, and SUTrack~\cite{sutrack}), trackers with cross-frame memory or context propagation (ODTrack~\cite{zheng2024odtrack}, SPMTrack~\cite{spmtrack}, and MCITrack~\cite{mcitrack}), an autoregressive sequence generation tracker (SeqTrack~\cite{chen2023seqtrack}), and a reinforcement learning box regression tracker (RELO~\cite{relo}). We compare WhereTrack on each base tracker against two search region adaptation baselines: SRRT~\cite{srrt}, which dynamically scales the search window size according to the tracking state, and LSTM~\cite{shahbazi2018motion}, which regresses a single next window center from historical trajectories. In all experiments, PointFlow uses four-step Euler integration to generate $K=4$ candidate centers per frame from the historical center sequence $h_t$. Policy combines the $7\times 7\times C$ local visual feature $s_v$, extracted from the search feature map $F$ via ROI-Align~\cite{he2017mask}, with the motion feature $h$ to evaluate all candidates and the default center $a_0$, determining the next search center $O_{t+1}$ via the switching threshold $\delta$ in Eq.~(\ref{eq:gate_rule}).

\noindent\textbf{Evaluation metrics.}
On FMD, NfS~\cite{galoogahi2017need}, LaSOT~\cite{fan2019lasot}, LaSOT-ext~\cite{fan2021lasot}, and TrackingNet~\cite{muller2018trackingnet}, we follow the one-pass evaluation (OPE) protocol~\cite{wu2015object} and report the area under the success curve (AUC), normalized precision ($\mathrm{P}_{\mathrm{norm}}$), and precision (P). On GOT-10k~\cite{huang2019got}, we follow the official protocol and report the average overlap (AO) and success rates at overlap thresholds of 0.5 and 0.75 ($\mathrm{SR}_{0.5}$ and $\mathrm{SR}_{0.75}$). To account for sampling stochasticity in PointFlow, all reported results involving WhereTrack are averaged over 10 independent runs throughout our experiments.

\subsection{Main Results}

\noindent\textbf{Results on fast-motion benchmarks.}
Table~\ref{tab:main_results} presents the quantitative results of the eight base trackers equipped with different search region adaptation strategies. On FMD and NfS~\cite{galoogahi2017need}, which feature frequent fast motion, integrating WhereTrack improves the average AUC across the eight base trackers by \textbf{+5.05\%} (from 49.56\% to 54.61\%) and \textbf{+0.90\%} (from 68.86\% to 69.76\%), respectively. In particular, MCITrack~\cite{mcitrack} achieves the largest AUC gain of \textbf{+7.28\%} (from 42.81\% to 50.09\%) on FMD, while ODTrack~\cite{zheng2024odtrack} improves by \textbf{+6.29\%} to reach the highest overall AUC of 61.22\%.

\begin{table}[t]
\centering
\footnotesize
\caption{Attribute-specific performance (AUC, \%) on FMD using ODTrack~\cite{zheng2024odtrack} as the baseline.}
\label{tab:attribute_results}
\setlength{\tabcolsep}{4pt}
\renewcommand{\arraystretch}{1.10}
\begin{tabular*}{\columnwidth}{@{\extracolsep{\fill}}lccc}
\toprule
Attribute & ODTrack & +WhereTrack & $\Delta$AUC \\
\midrule
Fast Motion (FM) & 53.04 & \textbf{60.61} & \textbf{+7.57} \\
Motion Blur (MB) & 52.33 & \textbf{58.35} & \textbf{+6.02} \\
Camera Motion (CM) & 65.61 & \textbf{69.02} & \textbf{+3.41} \\
Small Target (ST) & 48.99 & \textbf{56.09} & \textbf{+7.10} \\
Low Resolution (LR) & 47.41 & \textbf{55.11} & \textbf{+7.70} \\
Occlusion (OCC) & 54.36 & \textbf{61.79} & \textbf{+7.43} \\
Aspect Ratio Change (ARC) & 56.59 & \textbf{61.53} & \textbf{+4.94} \\
Scale Variation (SV) & 57.41 & \textbf{62.08} & \textbf{+4.67} \\
\bottomrule
\end{tabular*}
\vspace{-6pt}
\end{table}

Comparing different adaptation strategies on the same base tracker reveals distinct behaviors. First, although SRRT~\cite{srrt} enlarges the search window to keep high-speed targets within view (bringing gains on SeqTrack, MCITrack, and SUTrack on FMD), expanding the cropping area introduces excessive background distractors and alters the target-to-window scale ratio, which severely degrades localization accuracy on trackers such as SPMTrack (\textbf{$-$14.27\%} on FMD) and LoRAT (\textbf{$-$19.92\%} on FMD), as well as across all eight trackers on NfS. Second, while LSTM~\cite{shahbazi2018motion} preserves the original window size and predicts the next search center from historical coordinates, deterministic single-point regression only extrapolates along the historical inertia direction, making it vulnerable to sudden stops or sharp non-linear turns. In contrast, WhereTrack generates diverse candidate centers via PointFlow and evaluates them with local visual cues via Policy, consistently outperforming LSTM across all eight base trackers on FMD and achieving the best overall performance across both fast-motion datasets. Notably, even for ODTrack~\cite{zheng2024odtrack}, SPMTrack~\cite{spmtrack}, and MCITrack~\cite{mcitrack}, which already incorporate cross-frame memory or temporal context propagation, integrating WhereTrack still yields substantial AUC gains of \textbf{+6.29\%}, \textbf{+6.01\%}, and \textbf{+7.28\%} on FMD, respectively. This confirms that actively shifting the search window center before image cropping is essential to prevent fast-moving targets from exiting the field of view.

\noindent\textbf{Results on general benchmarks.}
Across the four general benchmarks LaSOT~\cite{fan2019lasot}, LaSOT-ext~\cite{fan2021lasot}, TrackingNet~\cite{muller2018trackingnet}, and GOT-10k~\cite{huang2019got}, integrating WhereTrack yields an average AUC gain of \textbf{+1.48\%} on LaSOT-ext (up to \textbf{+2.97\%} on SPMTrack~\cite{spmtrack}) and \textbf{+0.60\%} on LaSOT (up to \textbf{+0.86\%} on SUTrack~\cite{sutrack}), while remaining comparable to the baselines on TrackingNet (\textbf{+0.04\%} AUC) and GOT-10k (\textbf{+0.18\%} AO, with minor fluctuations on individual trackers such as \textbf{$-$0.53\%} on SPMTrack). We put more analysis in the supplementary material.


\begin{figure}[t]
\centering
\includegraphics[width=0.98\columnwidth]{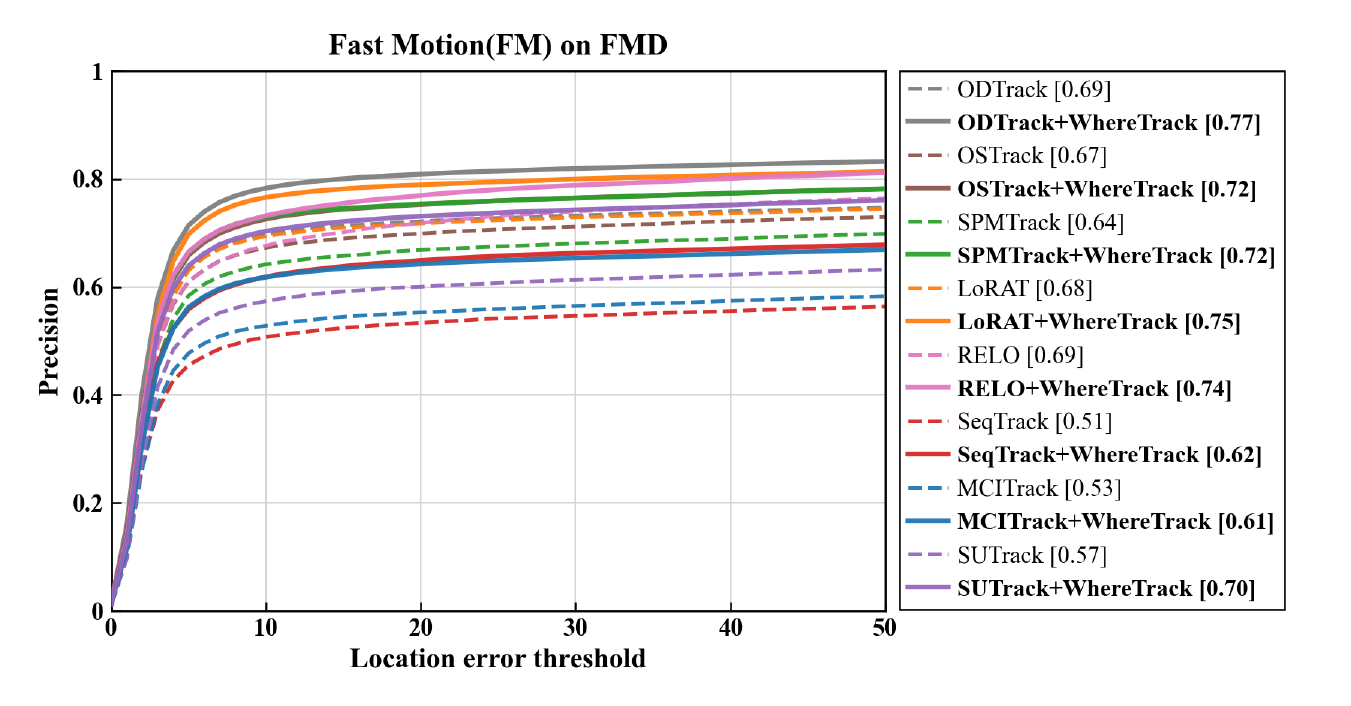}
\vspace{-6pt}
\caption{Attribute-based precision plots on the Fast Motion (FM) subset of FMD across eight base trackers with and without WhereTrack. Solid and dashed curves denote \textbf{+WhereTrack} and the original baselines, respectively.}
\label{fig:fm_precision}
\vspace{-6pt}
\end{figure}

\noindent\textbf{Attribute analysis on FMD.}
Table~\ref{tab:attribute_results} reports the attribute-wise AUC of ODTrack across eight challenging subsets of FMD, while Fig.~\ref{fig:fm_precision} presents the precision curves of eight base trackers on the Fast Motion (FM) subset. WhereTrack improves ODTrack's AUC by \textbf{+7.57\%} on FM, \textbf{+6.02\%} on Motion Blur (MB), and \textbf{+3.41\%} on Camera Motion (CM). All eight trackers consistently benefit from WhereTrack under fast motion, demonstrating its generalizability. By predicting search centers from historical trajectories, PointFlow reduces reliance on degraded visual appearances. WhereTrack also achieves substantial gains on Small Target (ST), Low Resolution (LR), and Occlusion (OCC), with AUC improvements of \textbf{+7.10\%}, \textbf{+7.70\%}, and \textbf{+7.43\%}, respectively. These gains are particularly valuable for small targets, which are more prone to leaving the search window. Further analysis is provided in the supplementary material.
\begin{figure*}[t]
\centering
\includegraphics[width=0.98\textwidth]{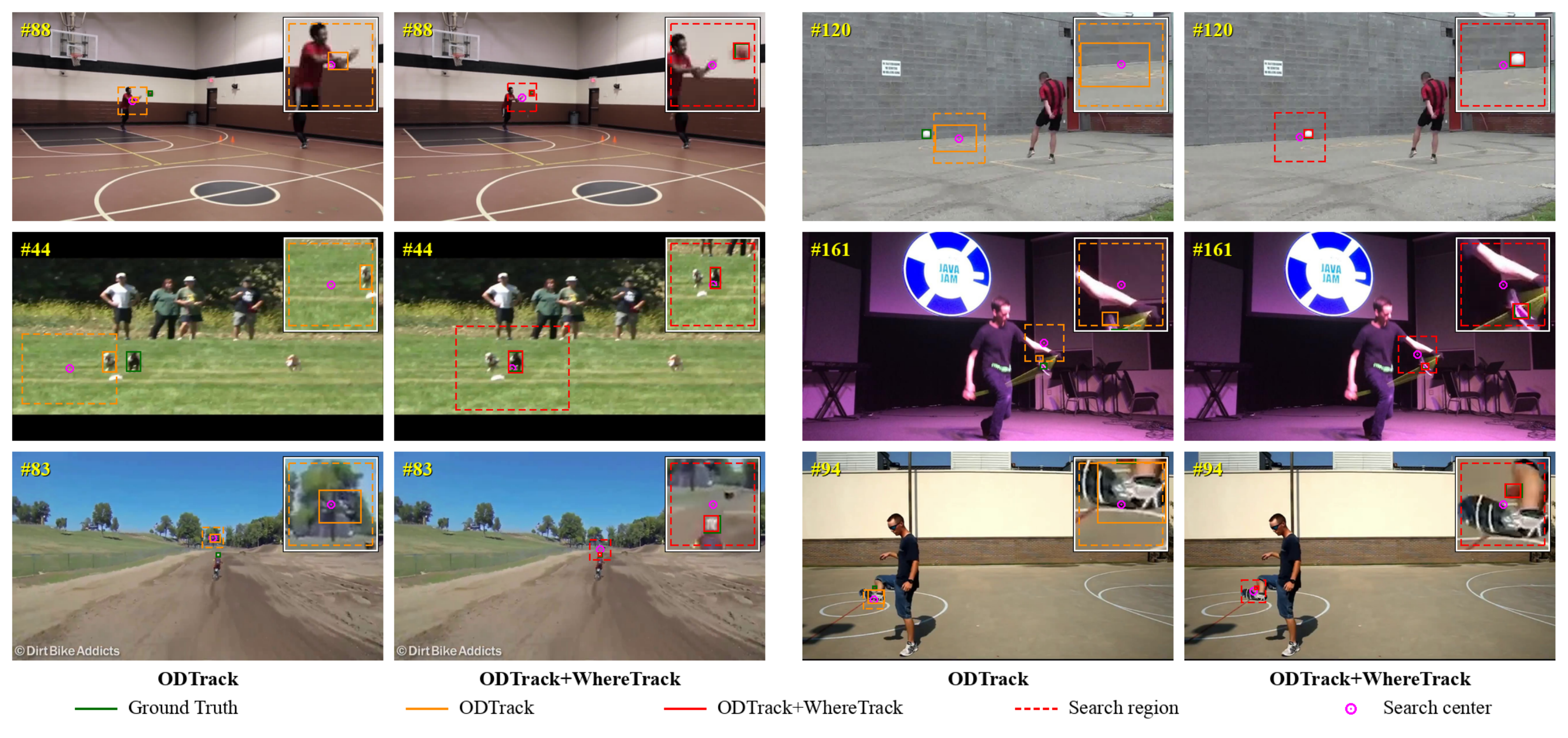}
\vspace{-6pt}
\caption{Qualitative comparisons between ODTrack and ODTrack+WhereTrack on challenging fast-motion sequences from FMD, with zoomed-in search regions shown in the top-right insets. Best viewed by zooming in.}
\label{fig:qualitative}
\vspace{-6pt}
\end{figure*}

\noindent\textbf{Qualitative analysis.}
Fig.~\ref{fig:qualitative} visually compares ODTrack and ODTrack+WhereTrack across six challenging fast-motion sequences on FMD, covering rapid ball flights in indoor sports and billiards as well as abrupt hand and object movements. As highlighted by the zoomed-in search regions in the top-right insets, the static search center in baseline ODTrack lags behind the target's high-speed trajectory, leaving the ground-truth target (green box) near the boundary or completely outside the cropped search region (orange dashed box) and resulting in localization drift. By contrast, ODTrack+WhereTrack proactively shifts the search center ahead of the rapid motion so that the red search region accurately re-centers on the target, enabling tight bounding box predictions (red box) aligned with the ground truth.

\subsection{Ablation Study and Efficiency Analysis}
Using ODTrack~\cite{zheng2024odtrack} as the baseline tracker on FMD, we conduct ablation studies on PointFlow and Policy (Table~\ref{tab:ablation}) and evaluate model parameters, computational overhead, and inference speed (Table~\ref{tab:efficiency}).

\begin{table}[t]
\centering
\footnotesize
\caption{Ablation study of PointFlow and Policy on FMD using ODTrack~\cite{zheng2024odtrack} as the baseline.}
\label{tab:ablation}
\setlength{\tabcolsep}{6.0pt}
\renewcommand{\arraystretch}{1.15}
\aboverulesep=0ex
\belowrulesep=0ex
\begin{tabular}{c|cc|ccc}
\toprule[1.2pt]
\# & \textbf{PointFlow} & \textbf{Policy} & AUC (\%) & $\mathrm{P}_{\mathrm{norm}}$ (\%) & P (\%) \\
\midrule[0.6pt]
\textbf{1} & \ding{55} & \ding{55} & 54.93 & 66.70 & 72.44 \\
\textbf{2} & \ding{51} & Random    & 49.98 & 60.42 & 66.17 \\
\textbf{3} & \ding{51} & Mean      & 60.53 & 73.65 & 79.95 \\
\textbf{4} & \ding{51} & \ding{51} & \textbf{61.22} & \textbf{74.40} & \textbf{80.94} \\
\bottomrule[1.2pt]
\end{tabular}
\vspace{-6pt}
\end{table}

\begin{table}[t]
\centering
\footnotesize
\caption{Comparison of model parameters, computational cost (FLOPs), and inference speed (FPS) based on ODTrack~\cite{zheng2024odtrack}.}
\label{tab:efficiency}
\setlength{\tabcolsep}{2.5pt}
\renewcommand{\arraystretch}{1.10}
\begin{tabular*}{\columnwidth}{@{\extracolsep{\fill}}lccc}
\toprule
Method & Params (M) & FLOPs (G) & FPS \\
\midrule
ODTrack & \textbf{92.83} & \textbf{73.10} & \textbf{67} \\
+ SRRT~\cite{srrt} & 101.69 & 85.03 & 35 \\
+ LSTM~\cite{shahbazi2018motion} & 94.09 & 73.10 & 63 \\
+ PointFlow (w/o Policy) & 109.97 & 74.05 & 56 \\
\rowcolor{officegrey}
\textbf{+ WhereTrack (PointFlow + Policy)} & 113.80 & 74.15 & 54 \\
\bottomrule
\end{tabular*}
\vspace{-6pt}
\end{table}

\noindent\textbf{Ablation analysis.}
Table~\ref{tab:ablation} evaluates the individual contributions of PointFlow and Policy on FMD. Compared with the baseline ODTrack (\textbf{\#1}, 54.93\% AUC), generating candidate centers via PointFlow but randomly selecting one without visual guidance (\textbf{\#2}, Random) degrades AUC by \textbf{4.95\%} to 49.98\%, as unguided random shifts frequently displace the search window away from the true motion trajectory. Taking the mean of the generated candidates (\textbf{\#3}, Mean) substantially boosts AUC to 60.53\% (\textbf{+5.60\%} over the baseline), demonstrating that the trajectory distribution learned by PointFlow effectively captures rapid target displacement. However, averaging multiple candidate modes collapses diverse directional hypotheses toward their centroid, which remains suboptimal when targets undergo abrupt stops or sharp non-linear turns. By incorporating Policy (\textbf{\#4}) to evaluate candidates with local visual and motion features, WhereTrack further improves performance to \textbf{61.22\%} AUC, \textbf{74.40\%} $\mathrm{P}_{\mathrm{norm}}$, and \textbf{80.94\%} P, outperforming the Mean strategy by \textbf{+0.69\%} AUC and \textbf{+0.99\%} P, and surpassing the baseline by \textbf{+6.29\%} AUC.

\noindent\textbf{Efficiency analysis.}
Table~\ref{tab:efficiency} compares the model parameters (Params), per-frame computational cost (FLOPs), and inference speed (FPS) on ODTrack~\cite{zheng2024odtrack}. Since PointFlow operates purely on low-dimensional historical coordinates without image feature computation, and Policy reuses the search feature map $F$ already extracted by the base tracker to run a lightweight MLP on the $7\times 7\times C$ ROI feature and $K+1=5$ candidate centers, the full WhereTrack adds only \textbf{20.97} M parameters (17.14 M from PointFlow and 3.83 M from Policy) and \textbf{1.05} G FLOPs (+1.4\% over the 73.10 G baseline). In contrast, SRRT~\cite{srrt} incurs an additional 11.93 G FLOPs and drops the speed to 35 FPS. Regarding inference speed, integrating WhereTrack reduces the frame rate from \textbf{67} FPS to \textbf{54} FPS (still running well above real-time requirements and $1.54\times$ faster than SRRT). This latency mainly stems from the four-step iterative Euler integration in PointFlow: although each step requires negligible FLOPs, sequential execution per frame introduces minor serial overhead. Overall, WhereTrack delivers substantial accuracy gains under fast motion with only 1.4\% additional FLOPs and real-time inference speed.
\section{Conclusion}

In this paper, we propose WhereTrack, a dynamic search center prediction framework that mitigates target loss under fast motion by repositioning the search window before cropping. It combines PointFlow-based candidate generation with policy-guided selection and integrates seamlessly into existing trackers without architectural modifications. We also introduce FMD, a benchmark of 178 sequences across 48 categories with 81.5\% fast-motion coverage. Experiments with eight trackers across six benchmarks demonstrate its effectiveness and generalizability.

\textbf{Limitations} WhereTrack remains dependent on the tracking capability of the base tracker. Future work will explore improving base tracker robustness and jointly optimizing tracking and search window prediction.


\section*{LLM Usage Statement}
Large Language Models (LLMs) were used solely for language editing, grammar checking, and stylistic polishing of author-written drafts. All research ideas, theoretical formulations, dataset construction and annotations, model implementations, figures, and experimental evaluations were conceived, executed, and verified exclusively by the authors. The authors have thoroughly reviewed all edited text to ensure technical accuracy, citation validity, and compliance with academic integrity standards, and take full responsibility for the entire submission.

{
    \small
    \bibliographystyle{ieeenat_fullname}
    \bibliography{main}
}

\end{document}